# Entropy in Semantic Memory Navigation in Blind and Sighted Individuals: The Effect of Visual Experience

**Felipe D. Toro-Hernández (ftoroher@gmail.com)[1], Rodrigo Lagos[2] & Sergio E. Chaigneau[2]**
[1]Graduate Program in Neuroscience and Cognition, Federal University of ABC (UFABC)
[2]Center for Social and Cognitive Neuroscience, Universidad Adolfo Ibáñez, Santiago, Chile

**Abstract**

Embodied accounts of semantic memory highlight the role of sensorimotor systems in acquiring and storing knowledge. Congenitally blind populations offer a critical test bed for these assumptions, providing an opportunity to assess whether conceptual grounding requires visual experience. In this study, we assessed semantic memory navigation differences between blind and sighted individuals using a property listing task with concrete and abstract concepts. We computed semantic entropy, an embedding-based natural language processing metric that captures the predictability of retrieval. Generalized linear mixed models revealed distinct navigation patterns across groups: while sighted individuals showed higher entropy for abstract than concrete concepts, blind participants did not. Instead, blind individuals exhibited higher entropy for visually salient concrete concepts (e.g., penguin). These results underscore the role of visual experience in the organization and dynamic navigation of semantic memory.

## Introduction

Semantic memory is a dynamic store of conceptual knowledge about the world that allows us to categorize and understand reality (Binder & Desai, 2011). When we think about concepts such as “dog” or “love,” we may evoke a range of associated experiences (e.g., playing with dogs, falling in love in the summer). In cognitive neuroscience, debates about how semantic knowledge is represented have long oscillated between embodied and disembodied accounts (Calzavarini, 2025; Mahon & Caramazza, 2008). Embodied approaches emphasize a central role for sensorimotor systems in the evocation and retrieval of semantic content (e.g., a dog’s shape, color, or texture; Barsalou et al., 2008; Barsalou, 2023; Kiefer & Pulvermuller, 2012), whereas more disembodied views posit amodal linguistic–semantic networks that support knowledge about the world (Patterson, Nestor & Rogers, 2007; Pylyshyn, 1984). Importantly, this debate often focuses on which features are most relevant for semantic representation, and particularly the extent to which sensorimotor systems contribute to the storage and access of meaning (Meteyard, et al., 2012). Contemporary grounded cognition work underscores the need to specify the degree of sensorimotor involvement, as empirical findings remain mixed and have made it difficult to fully reconcile the evidence with a single, unified account (Binder & Desai, 2011; Meteyard, et al., 2012).

To address these questions, studies in diverse populations have been crucial for clarifying the role of sensorimotor systems in semantic memory. Research examining groups with altered neural or sensory pathways that are thought to support conceptual representations can provide strong tests of competing accounts (as Parkinson’s disease, see Boulenger, et al., 2008; Birba, et al., 2017; Fernandino, et al., 2013). Furthermore, a substantial body of evidence comes from comparisons between blind and sighted individuals (Bottini, et al., 2020; Lenci, et al., 2013; Mamus, et al., 2025). From strong embodied accounts, blindness should lead to systematic differences in semantic representations, given the absence of visual experience and visual traces that might otherwise contribute to conceptual knowledge (Connolly, Gleitman & Thompson-Schill, 2007). In principle, this could affect the retrieval of semantic information, especially for concepts and features that are typically vision-dependent (e.g., “it’s shiny” for SUN, or “it’s red” for APPLE).

Interestingly, the literature has often shown that, despite the loss of visual sensory traces, blind individuals can acquire and use many aspects of language and conceptual knowledge in ways that are broadly comparable to sighted individuals (Landau & Gleitman, 1985). This finding has been interpreted in terms of the central role of linguistic and social input, in which semantic knowledge can be learned and transmitted through contextual regularities in language use, communication, and shared cultural practices, even in the absence of direct visual experience (Canessa et al., 2021). Crucially, evidence from neuroplasticity research suggests substantial functional reorganization in blindness, supporting the recruitment of visual cortical regions for nonvisual cognition. In line with this, neuroimaging studies have repeatedly reported that occipital cortex in blind participants is engaged during linguistic–semantic processing (Bedny et al., 2011; Bedny, Richardson & Saxe, 2015) and can also contribute to tactile reading, including Braille (Sadato et al., 1996; Sadato, 2005).

Despite generally preserved language abilities, studies that zoom in on specific semantic domains have reported selective differences in concept knowledge and categorization in blindness. For example, Saysani, Corballis & Corballis (2021) found broadly similar performance in semantic categorization, with a notable exception for color, a highly salient visual property. Whereas sighted participants’ color categorizations were relatively consistent, blind participants showed substantially greater variability, in ways that appeared to track differences in linguistic experience with color terms. Converging neuroimaging evidence suggests both shared and group-specific mechanisms. Wang et al. (2020) reported overlapping regions engaged by color language in blind and sighted participants, alongside additional activations specific to sighted individuals, consistent with a contribution of visual experience to the neural representation of color. Similarly, Bottini et al. (2020) showed that even when behavioral performance (e.g., reaction times) was comparable across groups, sighted participants more strongly engaged visual cortical regions during judgments in highly visual domains such as color, indicating that similar behavior can be supported by different neural routes. In line with this, Connolly et al. (2007) found in implicit similarity judgments that color was a diagnostic dimension for sighted but not blind participants when evaluating items such as foods and vegetables.

This particular profile in blind individuals (showing responses similar to those of sighted participants despite differences in conceptual content) is especially informative for the question of *when* sensorimotor representations are necessary for conceptual processing in the embodiment debate (Meteyard, et al., 2012; Ostarek & Bottini, 2021). Because visual sensory input is unavailable, yet participants can still perform successfully on semantic tasks involving color, this may suggest that sensorimotor representations are not a *sine qua non* condition for semantic processing to occur. At the same time, the observed differences in concept comprehension indicate that, even if they are not essential for forming and using semantic representations, sensorimotor systems can still contribute to how semantic knowledge is acquired, recovered, and stored. As Ostarek & Bottini, (2021) argues, sensorimotor systems may have functional importance for semantic processing, without being strictly necessary for semantic representations to be successfully accessed and used.

Studies using (dis)similarity ratings, reaction times, or property listing measures offer different views on blind semantic processing. For instance, Lenci et al. (2013) drew on Wu and Barsalou's (2009) taxonomy to examine qualitative differences in semantic processing between blind and sighted individuals. They found that blind participants tended to produce fewer perceptual properties than sighted participants, suggesting that relative to sighted, blind participants may rely less on perceptual feature description and more on contextual/encyclopedic ones (e.g., associated objects/events; Lenci et al. 2013). Building on the same dataset, Canessa et al. (2021) analyzed mathematical regularities in the time course of the property listing task to characterize the dynamics of semantic memory retrieval. Their results indicated that, whereas sighted participants showed differences between abstract and concrete concepts, blind participants displayed more similar retrieval dynamics across these concept types, suggesting a more comparable search profile for concrete and abstract knowledge.

According to contemporary accounts of abstract concept acquisition, differences between abstract and concrete concepts are often attributed to the former's stronger reliance on linguistic and contextual information and their relative lack of stable perceptual features during learning (Borghi et al., 2019). More specifically, converging evidence suggests that abstract and concrete concepts are acquired through partially distinct developmental pathways. Abstract concept learning appears to require more advanced cognitive and metacognitive resources and is strongly shaped by social interaction and linguistic mediation, a pattern that is consistent with evidence derived from word association paradigms and related semantic tasks (see Borghi et al., 2019; Borghi, 2023). Because blind individuals typically acquire knowledge about many concrete categories primarily through linguistic input rather than visual sensory cues, one plausible hypothesis is that concrete and abstract concepts may overlap more in their retrieval processes for blind individuals than for sighted individuals. However, we still know relatively little about how semantic retrieval unfolds over time in blindness, and whether blindness compresses the usual abstract–concrete differences in retrieval dynamics.

In this study, we aimed to assess differences between concrete (e.g., animals, vegetables, fruits) and abstract (e.g., joy, jealousy, democracy) concepts by using natural language processing (NLP) methods to capture the step-by-step granularity of semantic memory navigation in congenitally blind and matched sighted individuals, using Lenci et al. (2013)'s original property-listing sequences. Traditional approaches to semantic search often rely on clustering and switching metrics to characterize optimal foraging strategies, in which participants balance exploration and exploitation (Hills et al., 2015; Troyer & Moscovitch, 1997). In addition, semantic coding approaches are frequently used, in which human coders judge each produced property and assign it to a category label (e.g., grouping "the cops" and "security forces" under a code such as "police"). While informative, these approaches can be time-consuming and difficult to scale, and to standardize across protocols (Bolognesi, Pilgram & Van den Heerik, 2017; Ramos, et al., 2025).

Here, we use an embedding-based analysis that represents property production as movement through semantic space (Toro-Hernández et al., 2026). This approach complements traditional semantic memory analyses by quantifying continuous semantic change along the retrieval sequence, capturing fine-grained variation from one step to the next while minimizing the need for manual coding or preprocessing implemented in NLP procedures (Corcoran, et al., 2018). Embedding-based methods have been widely used to characterize semantic structure across concepts and have also been applied in cognitive and clinical research to quantify group differences in language-derived semantic representations (Toro-Hernández et al., 2026; Fraser, et al., 2019; Linz, et al., 2017; Nour, et al., 2025; Palominos, et al., 2024).

Thus, we use semantic entropy to quantify patterns of unpredictability in blind and sighted individuals who performed a property listing task. Entropy captures semantic variation across successive steps in the listing process, providing a window into the dynamics of semantic memory retrieval. For sighted participants, we expect abstract concepts to exhibit more entropic search dynamics because they are typically harder to define and tend to elicit greater variability across speakers, partly due to the absence of a clear and widely shared perceptual referent and greater reliance on contextual and linguistic information (Borghi et al., 2019). Crucially, if sensorimotor experience is a primary driver of the concrete/abstract distinction, we hypothesize a compression of this entropy gap in blind individuals. Specifically, we predict that blind participants will show more comparable retrieval dynamics across both domains, as their knowledge of concrete categories is likewise mediated primarily through linguistic-contextual pathways, rendering their search profile more similar to that of abstract concepts.

## Methods

### Dataset: participants, task, and stimuli

In this study, we analyzed the semantic feature norms from the BLIND dataset published by Lenci et al. (2013). The dataset consists of 48 native Italian-speaking participants, with 22 congenitally blind and 26 sighted. The groups were matched as closely as possible in age, education, and sociocultural background. Mean age was 47.2 years (SD = 16.5) in the blind group and 45.1 years (SD = 16.8) in the sighted group. All participants reported no history of neurological or language disorders.

Participants performed a property listing task in which they heard words in randomized order and were asked to orally describe its meaning for 1 minute. To reduce fixation on specific concepts, they were instructed “not to hurry,” despite the time limit. These data were collected to investigate the nature of semantic representations both with grounded visual cues (in sighted participants) and without them (in congenitally blind participants). To ensure minimal data preprocessing, all raw responses were included in the embedding analyses, including all words (content words, stop words) as they were transcribed by the authors in the original work.

The full dataset comprises property listings for 50 noun concepts and 20 verb concepts spanning both concrete and abstract semantic categories. In the present study, we focused on a subset of noun concepts that enabled a principled comparison between concrete and abstract categories. Specifically, we selected concrete and abstract categories from the BLIND dataset (BIRD, MAMMAL, FRUIT, VEGETABLE, EMOTION, and IDEAL) to assess group differences in semantic organization across categories. The BIRD category included penguin, canary, swan, seagull, and crow; MAMMAL included cat, dog, horse, zebra, and giraffe; FRUIT included banana, cherry, apple, pineapple, and kiwi; VEGETABLE included potato, tomato, carrot, lettuce, and eggplant; EMOTION included cheerfulness, passion, pain, jealousy, and worry; and IDEAL included democracy, justice, religion, freedom, and friendship. As reported in the original work, there were no differences in the total number of features produced per subjects across groups.

### NLP analysis

We performed natural language processing (NLP) analyses over the raw data of the dataset, using a transformer-based multilingual embedding model (Google Gemini Embedding; gemini-embedding—004). Specifically, we mapped the semantic representation of each feature produced in the property listing task to obtain a cumulative sequence of meaning across responses, following Toro-Hernández et al. (2026). Each feature was represented as a point in a multidimensional semantic space, and we modeled its sequential order dependence to capture the cumulative nature of semantic retrieval. This approach is motivated by evidence that, when individuals generate features in fluency-like tasks, multiple processes are engaged. Items must be temporarily maintained in working memory while new information is retrieved from semantic memory, and previously produced items must be inhibited to avoid repetitions (proactive interference; Baddeley, 2000).

To operationalize this idea, we treated each participant–concept response as an ordered stream of produced features of length N and mapped it to a trajectory in semantic space $X = (x_1, \ldots, x_N)$, where $x_t$ is the point associated with the t-th step in the sequence. We construct the embeddings cumulatively, such that each point summarizes the full prefix 1:t. Formally, letting $E(\cdot)$ denote the embedding model, we computed:

$$x_t = E(w_1, w_2 \ldots w_t), \quad t = 1, \ldots, N$$

where $w_t$ is the feature produced at step t. For example, in the sequence “dog”, “cat”, and “rat”, then $x_1$ encodes “dog”, $x_2$ encodes “dog cat”, and so on. Thus, at each step, the embedding represents the semantic content of the current property together with all previously produced properties for that concept and participant. This procedure yields a semantic trajectory through embedding space for each participant–concept pair. From each trajectory, we computed semantic entropy, a measure indexing unpredictability in semantic memory navigation over the listing process.

Entropy was calculated by first computing the cosine distance between successive cumulative embeddings, producing a sequence of semantic “jumps” of length $n = N - 1$. We then split these distances at their within-trajectory median θ, yielding a binary time series $b_t$ defined as:

$$b_t = 1 \text{ if } d_t \geq \theta;\ 0 \text{ otherwise } (d_t < \theta), \text{ for } t = 1, \ldots, n$$

Let p be the fraction of “high-jump” (or ones) steps along the trajectory:

$$p = (1/n)\, \Sigma_{t=1}^{n}\, b_t$$

Shannon entropy was then computed over the resulting binary sequence:

$$H = -p\log_2(p) - (1-p)\log_2(1-p)$$

Because entropy is computed from the proportion of high-jump steps (p) over the trajectory, it is comparable across trajectories of different lengths (Shannon, 1948). Higher entropy values indicate more variable and less predictable semantic exploration, whereas lower values indicate more constrained and regular semantic trajectories. In the example above, entropy would be lower because the sequence is more predictable. In contrast, if the sequence were “dog”, “shark”, “cat”, entropy would be higher because the greater variability reflects more uncertainty in the listing process.

In terms of semantic memory navigation, more variable search dynamics may reflect difficulty in maintaining a consistent search strategy, such that semantic shifts in content are less expected than in a more predictable sequence (Toro-Hernández et al., 2026).

## Statistical analysis

We performed Generalized Linear Mixed Model (GLMM) analyses to assess group differences across semantic categories while accounting for the repeated-measures structure of the data. First, we assessed overall differences between abstract and concrete categories, and then examined concept-specific differences to determine whether group effects were broadly distributed across semantic domains or driven by specific concepts. The model included fixed effects of group, semantic category, and their interaction, and random intercepts for participant and concept, aiming to control for individual differences in semantic search strategies as well as concept-specific variability. We used a Gamma distribution (log link) based on residuals diagnostics using R’s DHARMa package. Post-hoc pairwise comparisons were conducted on model-estimated marginal means using Tukey’s adjustment to control the family-wise error rate. All analyses were performed in R (v.4.3.1) using the glmmTMB (Brooks et al., 2017) and the emmeans packages. All plots, GLMMs, and Tukey outputs, and entropy values used for all analyses can be found here: https://osf.io/pdwk2.

# Results

## GLMM category analysis

After fitting the model, we observed category-dependent group differences in semantic entropy. Using BIRD and the blind group as reference levels, sighted participants showed lower entropy than blind participants in the BIRD category ($\beta = -0.283$, $SE = 0.118$, $z = -2.40$, $p = 0.017$). Importantly, the group difference varied by category, as indicated by significant interactions for EMOTION ($\beta = 0.435$, $SE = 0.0756$, $z = 5.75$, $p < 0.001$), IDEAL ($\beta = 0.362$, $SE = 0.0774$, $z = 4.68$, $p < 0.001$), MAMMAL ($\beta = 0.197$, $SE = 0.0754$, $z = 2.62$, $p = 0.009$), and VEGETABLE ($\beta = 0.234$, $SE = 0.0748$, $z = 3.13$, $p = 0.002$), whereas the interaction for FRUIT was not reliable ($p = 0.178$). This pattern indicates that while sighted participants tended to show lower entropy than blind participants in several concrete categories, the difference reversed for abstract concepts, most prominently for emotion, where sighted participants showed higher entropy.

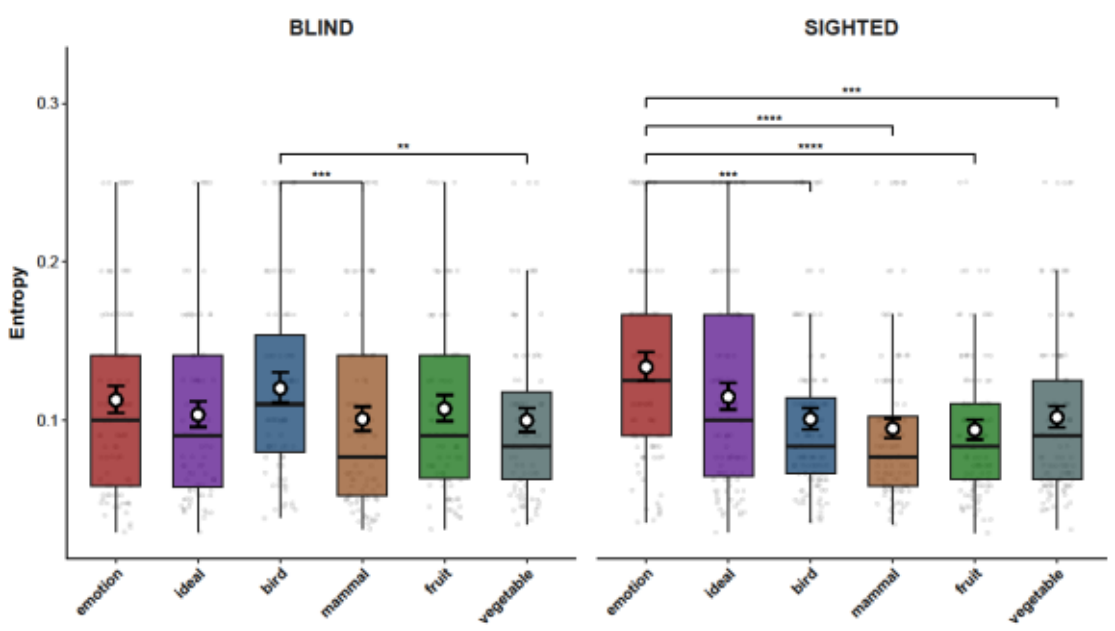

**Figure 1.** Entropy by semantic category in congenitally blind and sighted participants. Points show individual observations; boxes show medians and interquartile ranges. Brackets denote Tukey-adjusted pairwise contrasts within each group. * p < .05, ** p < .01, *** p < .001, **** p < .0001.

Tukey pairwise contrasts showed that, in the blind group, only a small number of contrasts survived correction, with BIRD showing higher entropy than MAMMAL ($\Delta = 0.290$, $p = 0.009$) and VEGETABLE ($\Delta = 0.260$, $p = 0.027$), and a marginal trend relative to ideal ($\Delta = 0.235$, $p = 0.067$). Notably, although not significant, this pattern suggests higher entropy for BIRD than for IDEAL in the blind group. In contrast, the sighted group showed robust category differentiation driven by EMOTION, which had higher entropy than BIRD ($\Delta = 0.274$, $p = 0.008$), FRUIT ($\Delta = 0.348$, $p < 0.001$), MAMMAL ($\Delta = 0.366$, $p < 0.001$), and VEGETABLE ($\Delta = 0.300$, $p = 0.002$), with no reliable difference between the two abstract categories (EMOTION vs IDEAL; $p = 0.465$). Overall, the results support a comparatively flatter entropy profile in blind participants and a more differentiated profile across categories in sighted participants, with elevated entropy for EMOTION concepts. Moreover, sighted participants tended to show higher entropy for abstract concepts, whereas the blind group showed the opposite (non-significant) tendency (see Figure 1).

## GLMM concept analysis

To examine concept differences within each group, we conducted Tukey-adjusted pairwise contrasts. Given the large number of possible comparisons, we summarize only the contrasts that survived correction, grouped by concept clusters that showed consistent patterns.

In the blind group, significant contrasts were concentrated in comparisons involving a small set of concepts, suggesting localized (rather than global) differentiation. First, cheerfulness showed higher entropy than several MAMMAL and VEGETABLE concepts, including dog ($\Delta = 0.559$, $p = 0.002$), horse ($\Delta = 0.474$, $p = 0.044$), cat ($\Delta = 0.550$, $p = 0.025$), and potato ($\Delta = 0.496$, $p = 0.014$). Second, several MAMMAL and BIRD related contrasts were robust, with dog showing lower entropy than swan ($\Delta = -0.580$, $p = 0.008$), crow ($\Delta = -0.594$, $p < .001$), giraffe ($\Delta = -0.484$, $p = 0.006$), kiwi ($\Delta = -0.482$, $p = 0.003$), eggplant ($\Delta = -0.545$, $p < .001$), penguin ($\Delta = -0.682$, $p < .001$), and zebra ($\Delta = -0.596$, $p = 0.002$). Third, contrasts involving penguin were especially prominent, showing higher entropy than tomato ($\Delta = 0.581$, $p < .001$) and worry ($\Delta = 0.440$, $p = 0.007$), and potato showing lower entropy than penguin ($\Delta = -0.620$, $p < .001$). Additional significant contrasts indicated lower entropy for cat than penguin ($\Delta = -0.673$, $p < .001$) and zebra ($\Delta = -0.587$, $p = 0.019$), as well as lower entropy for justice than penguin ($\Delta = -0.494$, $p = 0.012$).

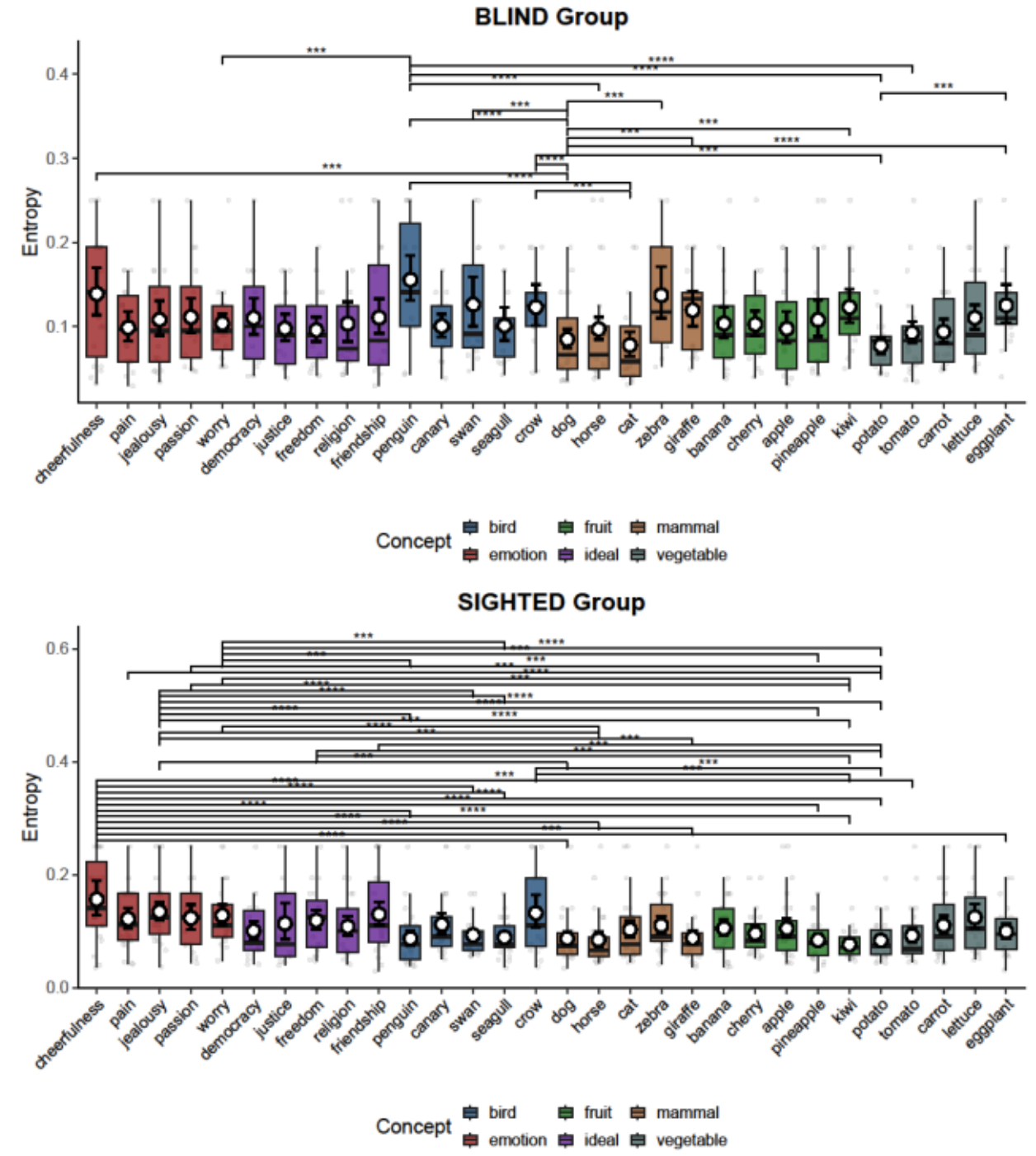

**Figure 2.** Entropy by concept in congenitally blind (top) and sighted (bottom) participants. Points show individual observations; boxes show medians and interquartile ranges. Blind participants show a smaller set of localized differences, whereas sighted participants show broader concept-level differentiation. $p < .05$, ** $p < .01$, *** $p < .001$, **** $p < .0001$.

In the sighted group, the number of significant contrasts was markedly larger, indicating stronger concept-level differentiation. Contrasts involving cheerfulness were especially frequent, showing higher entropy than multiple concrete concepts (e.g., pineapple, dog, horse, swan, seagull, giraffe, kiwi, eggplant, potato, penguin, and tomato; all Tukey-adjusted $p$'s $\leq 0.020$). In addition, a coherent set of contrasts emerged among EMOTION/IDEAL related concepts, with jealousy showing higher entropy than seagull ($\Delta = 0.399$, $p < .001$), giraffe ($\Delta = 0.424$, $p = 0.002$), kiwi ($\Delta = 0.467$, $p < .001$), potato ($\Delta = 0.467$, $p < .001$), penguin ($\Delta = 0.449$, $p < .001$), and tomato ($\Delta = 0.363$, $p = 0.016$). Finally, worry consistently differed from several concrete concepts, showing higher entropy than pineapple ($\Delta = -0.419$ for pineapple − worry, $p = 0.004$), dog ($\Delta = -0.379$, $p = 0.0295$), horse ($\Delta = -0.449$, $p = 0.006$), seagull ($\Delta = -0.381$, $p = 0.008$), kiwi ($\Delta = -0.450$, $p < .001$), potato ($\Delta = -0.449$, $p < .001$), and penguin ($\Delta = -0.431$, $p = 0.009$) (see Figure 2).

Together, these results indicate that the sighted group exhibited broader and more systematic concept-level differentiation, whereas the blind group showed fewer significant distinctions concentrated in a smaller subset of concepts (see Figure 2). Beyond contrasts involving cheerfulness, blind-group effects were largely confined to within-animal comparisons (e.g., dog vs. multiple BIRD/MAMMAL concepts; see Figure 2), whereas sighted-group contrasts were disproportionately driven by EMOTION/IDEAL concepts (most notably jealousy and worry) showing higher entropy than several concrete concepts (see Figure 2). Critically, the direction of significant abstract–concrete contrasts was consistent in the sighted group, with abstract concepts exhibiting higher entropy in all cases (especially jealousy and worry). In the blind group, however, this abstract–concrete pattern was less uniform. Although cheerfulness showed higher entropy than several concrete concepts, other contrasts indicated the opposite direction, with some abstract concepts showing lower entropy than concrete ones (e.g., penguin showing more entropy than justice and worry).

# Discussion

The present study investigated how congenitally blind and sighted individuals navigate semantic memory when producing properties for abstract and concrete concepts, using an embedding-based approach to capture the dynamics of semantic retrieval. By modeling property listing as movement through semantic space and quantifying its variability with semantic entropy, we aimed to go beyond static descriptions of semantic content and instead characterize how meaning unfolds over time (Toro-Hernández et al., 2026). Overall, the results reveal systematic differences in semantic memory navigation between blind and sighted participants, as well as between concrete and abstract concepts, providing new insights into the role of sensory experience in structuring semantic memory.

At a broad level, the results show that blind and sighted participants differ not so much in their ability to retrieve semantic knowledge, but in how that knowledge is organized and traversed. Sighted participants exhibited a more differentiated entropy profile across semantic categories and individual concepts, whereas blind participants showed a comparatively flatter profile, with fewer and more localized differences. This pattern suggests that semantic memory in blindness is organized in a way that reduces the contrast between concept types that are distinguished in sighted individuals, particularly along the abstract–concrete dimension. Importantly, this does not imply a deficit or impoverishment of semantic knowledge in blindness. Rather, it points to a reorganization of semantic space that reflects differences in the sources of information available during concept acquisition and use.

When considering category-level effects, the interaction between group and semantic category was especially informative. In sighted participants, abstract categories (most notably emotion concepts) consistently elicited higher entropy than concrete categories. This finding aligns with accounts emphasizing that abstract concepts lack stable perceptual referents and therefore rely more heavily on contextual, linguistic, and social information (Borghi et al., 2019; Borghi et al., 2023). The higher entropy observed for abstract concepts in sighted participants can thus be interpreted as reflecting more exploratory and less predictable semantic memory navigation, driven by the need to integrate heterogeneous sources of information when defining or describing these concepts. Concrete concepts, by contrast, benefit from perceptual grounding (particularly visual experience), which may constrain semantic search and lead to more regular and predictable retrieval trajectories (Barsalou, 2008; Binder & Desai, 2011).

In blind participants, however, this abstract–concrete contrast was markedly attenuated. Entropy levels were more similar across categories, and in some cases concrete concepts showed entropy values comparable to or even higher than those of abstract concepts. This pattern supports the idea that, in the absence of visual experience, concrete concepts may rely more heavily on linguistic, functional, and contextual information, much like abstract concepts (Canessa et al., 2021). As a result, the usual advantage that concrete concepts have in sighted individuals, which stems from rich and shared perceptual traces, may be reduced or absent. In line with this interpretation, the Lenci et al. (2013) norms report fewer perceptual properties in blind participants' listings, and Canessa et al. (2021) show that abstract–concrete differences are less pronounced across the unfolding of the listing process in the blind. More broadly, this convergence with behavioral and neuroimaging work suggests that comparable semantic performance can be supported by partially different representational routes when a dominant modality (vision) is unavailable (Connolly et al., 2007; Bottini et al., 2020).

At the concept level, the contrast between groups became even clearer. Sighted participants showed widespread concept-level differentiation, with several abstract concepts (such as jealousy and worry) reliably exhibiting higher entropy than many concrete items. This suggests that, for sighted individuals, abstract concepts are not only globally more variable but also may systematically differ from concrete concepts in their retrieval dynamics. In blind participants, by contrast, significant concept-level differences were fewer and often concentrated within specific domains, such as comparisons among animal concepts. Notably, some highly visual concrete items (e.g., penguin) showed elevated entropy in the blind group, in some cases exceeding abstract concepts. Because the stimulus set was designed to include concepts with salient visual attributes (Lenci et al., 2013), this pattern suggests that visually diagnostic concepts may be less uniformly organized without direct perceptual experience, yielding more variable semantic memory navigation.

Taken together, these results help reconcile two seemingly opposing strands of the literature on semantic cognition in blindness. On the one hand, numerous studies, dating back to Landau and Gleitman (1985), have shown that blind individuals possess largely intact and functional semantic knowledge. On the other hand, more fine-grained analyses have revealed selective differences in how specific types of information (e.g., perceptual features like color; Saysani et al., 2021) are represented and used. The present findings suggest that these differences may be most evident not in whether a concept can be accessed, but in how semantic memory is navigated during retrieval. Blind and sighted individuals appear to share a broadly similar semantic space, while differing in their internal geometry and the navigation through it, as a function of sensory experience.

The results are difficult to reconcile with strong versions of embodiment that treat sensorimotor representations as necessary for semantic processing, since blind participants clearly demonstrate coherent and systematic semantic retrieval. At the same time, the findings challenge strictly amodal views by showing that sensory experience influences the organization and dynamics of semantic memory. Our results suggest that sensorimotor systems are not a *sine qua non* condition for semantic representations, but they do play an important role in shaping how semantic knowledge is structured and navigated, particularly for concepts that are typically grounded in perception (Mahon & Caramazza, 2008; Ostarek & Bottini, 2021).

Several limitations of the present study should be acknowledged. First, although we relied on standard category distinctions between concrete and abstract concepts, concreteness itself is a graded and multidimensional property, and future work could benefit from more fine-grained measures that capture perceptual, functional, emotional, and social dimensions separately (Lynott et al., 2020; Troche et al., 2017). Second, the entropy measure, while informative, provides a relatively coarse index of semantic variability; complementary analyses of trajectory length, directional changes, or clustering structure in embedding space may yield additional insights. Finally, the data analyzed here were drawn from an existing dataset, and future studies could be designed specifically to manipulate sensory relevance, communicative goals, or task demands to more directly test causal hypotheses about semantic memory navigation.

In conclusion, by applying an embedding-based, dynamic analysis to property listing data, the present study provides novel evidence that blindness alters the organization and navigation of semantic memory without undermining semantic competence. Sighted individuals show greater differentiation between abstract and concrete concepts, reflected in higher entropy for abstract items, whereas blind individuals exhibit a compressed semantic landscape in which this distinction is reduced (and in some cases reversed). These findings contribute to a more nuanced understanding of how sensory experience shapes semantic cognition and highlight the value of trajectory-based approaches for uncovering subtle differences in the dynamics of meaning retrieval.

## Acknowledgments

During the preparation of this work the author(s) used LLMs to review code and write in specific parts of the manuscript. After using this tool/service, the author(s) reviewed and edited the content as needed and take(s) full responsibility for the content of the published article. Felipe Diego Toro-Hernández acknowledges financial support from the São Paulo Research Foundation (FAPESP), grant 2025/04243-2.